\pdfoutput=1

\documentclass[11pt]{article}

\usepackage[final]{ACL2023}
\usepackage{times}
\usepackage{latexsym}
\usepackage{booktabs} 
\usepackage{multirow}
\usepackage{siunitx} 
\usepackage{graphicx}
\usepackage[utf8]{inputenc}
\usepackage{hyperref}
\usepackage{babel,blindtext}
\usepackage[T1]{fontenc} 
\usepackage{amsmath,amssymb,mathptmx,amsfonts}
\usepackage{algorithmic}
\usepackage{textcomp}
\usepackage{makecell}
\usepackage[export]{adjustbox}
\usepackage{multirow}
\usepackage{multicol}
\usepackage{verbatim}
\usepackage[inline]{enumitem}
\usepackage{dirtytalk}
\usepackage{pgfplots}
\usepackage{tikz}
\usetikzlibrary{shapes,backgrounds}
\usetikzlibrary{shapes,arrows,calc}
\usepgfplotslibrary{statistics}
\usepackage{url}
\usepackage{float}
\usepackage{comment}
\usetikzlibrary{tikzmark}
\usetikzlibrary{calc}
\usepackage{caption}
\usepackage{subcaption}
\usepackage{alphalph}
\usepackage{balance}

\usepackage[scaled]{helvet} 
\usepackage{courier}  

\usepackage{times}
\usepackage{latexsym}
\usepackage{microtype}
\usepackage{tcolorbox}
\usepackage{url}
\usepackage{mathtools}
\usepackage{cleveref}
\usepackage[perpage]{footmisc}

\usepackage{lscape} 
\usepackage{microtype}
\usepackage{enumitem,kantlipsum}
\usepackage{adjustbox}
\usepackage{xcolor}
\usepackage{pifont}
\restylefloat{table}
\crefformat{section}{\S#2#1#3} 
\crefformat{subsection}{\S#2#1#3}
\title{Systematic Offensive Stereotyping (SOS) Bias in Language Models}

\author{Fatma Elsafoury \\
  Fraunhofer Research Institute  \\
  Weizenbaum Research Institute \\
  \texttt{fatma.elsafoury@fokus.fraunhofer.de} \\}

\begin{document}
\maketitle

\begin{abstract}
In this paper, we propose a new metric to measure the SOS bias in language models (LMs). Then, we validate the SOS bias and investigate the effectiveness of removing it. Finally, we investigate the impact of the SOS bias in LMs on their performance and fairness on hate speech detection. Our results suggest that all the inspected LMs are SOS biased. And that the SOS bias is reflective of the online hate experienced by marginalized identities. The results indicate that using debias methods from the literature worsens the SOS bias in LMs for some sensitive attributes and improves it for others. Finally, Our results suggest that the SOS bias in the inspected LMs has an impact on their fairness of hate speech detection. However, there is no strong evidence that the SOS bias has an impact on the performance of hate speech detection.
\end{abstract}

\section{Introduction}
Language models (LMs) are the new state-of-the-art models. They are being implemented in tools like search engines \citep{Zhu2023LargeLM} and content moderation \cite{elsafoury2021}. Research has shown that LMs, are socially biased \cite{nangia-etal-2020-crows, nadeem-etal-2021-stereoset}. However, the offensive stereotyping bias and toxicity in LMs are still understudied. \citet{nozza-etal-2021-honest} and \citet{nozza-etal-2022-measuring} demonstrate that LMs tend to generate hurtful content. \citet{ousidhoum-etal-2021-probing} demonstrate that when probed by words that describe different identity groups, English LMs generate words that are insulting 24\% of the time, in comparison to, stereotypical (13\%), confusing (25\%) and normal (38\%). These results suggest that LMs are toxic. However, they do not investigate whether LMs systematically give higher probability to profane content over non-profane one, when probed by different identity groups. 
 
On the other hand, \citet{elsafoury_sos_2022}, introduce systematic offensive stereotyping (SOS) bias and propose a method to measure it in 15 different static word embeddings. However, the SOS bias has not been measured in LMs. Moreover, \citet{elsafoury_sos_2022} measure and validate the SOS bias, but their investigation does not include removing the SOS bias or how effective the state-of-the-art debias methods are on removing the SOS bias. Additionally, \citet{elsafoury_sos_2022} investigate the impact of the SOS bias in static word embeddings on their performance on the task of hate speech detection, excluding the impact of the SOS bias on another critical aspect, which is the \textit{fairness} of hate speech detection. 

In this paper, we fill these research gaps by proposing a metric to measure the SOS bias in LMs (\cref{sec:measure-sos}). Then, we validate the proposed SOS bias metric by comparing it to social bias metrics (\cref{sec:sos-bias-wv-social-bias}). Additionally, we investigate how reflective the SOS bias is of the online hate experienced by marginalized groups (\cref{sec:sos-bias-vs-online-hate}). Thereafter, we investigate the effectiveness of removing the SOS bias using one of the state-of-the-art debias methods (\cref{sec:sos-bias-removal}). Finally, we investigate the impact of the SOS bias in the inspected LMs on their performance (\cref{sec:sos-impact-on-performance}) and their fairness (\cref{sec:sos-bias-impact-on-fairness}) of the downstream task of hate speech detection. \textbf{The main Contributions} of this work can be summarized as following: \textbf{(1)} We provide a comprehensive investigation of the systematic offensive stereotyping (SOS) bias in LMs. \textbf{(2)} We create a new dataset to measure the SOS bias in LMs. \textbf{(3)} We make the newly created dataset and the code used in this work available online\footnote{This link will be available upon acceptance.}

The \textbf{findings} of this work demonstrate that all the inspected LMs are SOS-biased. Our results suggest that for most of the examined sensitive attributes, the SOS bias scores are higher against marginalized identities. We demonstrate that the SOS bias in the inspected LMs is reflective of the hate and extremism that are experienced by marginalized groups online. However, we found no strong evidence that our proposed SOS bias metric reveals different information from social bias metrics. Our results suggest that removing SOS bias from LMs, using one of the state-of-the-art debias methods, improved the SOS bias scores in the inspected LMs regarding some sensitive attributes and worsened it for others. On the other hand, our results suggest that, for some bias metrics, removing the SOS bias significantly improved the social bias scores. Our results suggest that the SOS bias in LMs has an impact on their fairness on hate speech detection. However, there is no strong evidence that the SOS bias has an impact on the performance of the task of hate speech detection, which is inline with previous findings \cite{goldfarb-tarrant-etal-2021-intrinsic}.

\section{Background}
There are a few studies that investigate the toxic response of LMs when probed by different identity groups. \citet{nozza-etal-2021-honest} investigate the hurtful stereotypical text generated by LMs when prompted by template sentences that contain gendered identity words. The results show that when LMs prompted by female gendered identity, 9\% of the generated text referred to sexual promiscuity. \citet{nozza-etal-2022-measuring} follow similar approach to measure the hurtful stereotyping in sentence completion against the LGBTQIA+ community and found that 13\% of the time, LMs generated identity attacks. \citet{ousidhoum-etal-2021-probing} demonstrates that LMs are toxic against people from different communities, marginalized and non-marginalized. The authors use the masked language models (MLM) task to predict words corresponding to template sentences that contain words that describe different identity groups, and then the authors use a logistic regression model to label whether the predicted words are toxic or not. Finally, a human evaluation of a 100 of the predicted words was conducted, where the results indicate that only 24\% of the predicted words were insulting regardless of the context in the English LMs, 11\% in French LMs, and 12\% in Arabic LMs. 

Even though these studies show evidence that LMs are toxic, especially towards marginalized groups. They have some limitations. For example, they do not systematically measure the toxicity or offensive stereotyping in LMs. As they all rely on open text generation by the LMs, which could be normal, confusing, or hurtful \cite{ousidhoum-etal-2021-probing}. We speculate that this is the reason behind the low percentages of hurtful content that are being exposed by these studies. Moreover, \citet{ousidhoum-etal-2021-probing} use a logistic regression model to predict whether the generated text is toxic or not and then uses human annotators to verify the label. This method of measuring the toxicity in LMs is not sustainable, as human annotators could be biased \cite{shah-etal-2020-predictive} and not always accessible. \citet{elsafoury_sos_2022} introduce systemic offensive stereotyping (SOS) bias and propose a method to measure in static word embeddings. However, the SOS bias has not been yet investigated in LMs. 

On the other hand, there are various metrics in the literature to systematically measure social bias in LMs like \textit{SEAT} \cite{DBLP:conf/naacl/MayWBBR19}, \textit{CrowS-Pairs} \cite{nangia-etal-2020-crows}, and \textit{StereoSet} \cite{nadeem-etal-2021-stereoset}. In SEAT, the authors, inspired by the WEAT metric \cite{Caliskan2017} to measure bias in static word embeddings,  propose a method to measure social bias in LMs. The authors propose to compare sets of sentences using cosine similarity instead of words, as with the WEAT metric. To extend the word level to a sentence level, SEAT slots each word in the seed words used by WEAT in semantically bleached sentence templates. Similarly, \textit{CrowS-Pairs} and \textit{StereoSet} metrics are used to measure social bias in LMs. But instead of sentence templates, the authors use crowdsourced sentences and the MLM task to measure the social bias. The Crows-Pairs dataset contains 1,508 sentence pairs (stereotypical and non-stereotypical) and measures nine types of social bias. 
 The StereoSet dataset contains 8,498 sentence pairs to measure four types of social bias. 

To measure the systematic offensive stereotyping (SOS) bias in LMs, these metrics will fall short since the crowdsourced sentences contain socially stereotypical versus non-stereotypical sentences. In this paper, we mitigate the limitations of the current literature by proposing a method to measure SOS bias in LMs. We build on existing social bias metrics but instead of using stereotypical and non-stereotypical sentence-pairs, we create a new dataset of profane and non-profane sentence-pairs to measure SOS bias.

\section{Measure SOS bias in LMs}
\label{sec:measure-sos}
The SOS bias, as defined by \citet{DBLP:conf/acl/Elsafoury22} is, ``\textit{A systematic association in the word embeddings between profanity and marginalized groups of people}''. We use that definition to measure the SOS bias in three LMs, BERT-base-uncased \cite{DBLP:conf/naacl/DevlinCLT19}, RoBERTa-base \cite{roberta}, and ALBERT-base \cite{albert}. To measure the SOS bias in LMs, we draw inspiration from the CrowS-Pairs and metric \cite{nangia-etal-2020-crows} that uses the MLM task to measure social bias in LMs. We use the MLM task to measure how many times a LM associates a profane sentence versus a non-profane sentence with a certain identity groups.

\subsection{$SOS_{LM}$ bias dataset}
\label{sec:sos_bias_dataset}
To measure the SOS bias in LMs using the MLM task, we first need a dataset of sentence-pairs similar to the datasets used in the CrowS-Pairs and StereoSet metrics. But that dataset should have two key characteristics: 1) It should contain profane and non-profane sentence-pairs, instead of social stereotypical and non-stereotypical sentence-pairs. 2) It should also contain fine-grained information on the identity group present in the sentences, so we can measure the SOS bias against marginalized and non-marginalized identity groups. 

We create a dataset that contains these two characters using sentence templates, similar to work done in the literature \cite{DBLP:conf/naacl/MayWBBR19, ousidhoum-etal-2021-probing, nozza-etal-2022-measuring}. We use non-offensive identity (NOI) words to describe different identity groups \citet{elsafoury_sos_2022}. Because, as argued by \citet{antoniak2021}, NOI words are more coherent in describing different identity groups. Furthermore, we compiled a list of 78 NOI words to describe different marginalized and non-marginalized identities for six sensitive attributes: gender, race, sexual orientation, religion, disability\footnote{We use only words to describe disability because the words that describe the able-bodied are not commonly used and are not shared in the reviewed literature that is used to compile that list of words. }, and social class \cite{elsafoury_sos_2022, Dixon2018, sweeney2019, Borkan-etal-2019-naunced-metrics, nangia-etal-2020-crows, nadeem-etal-2021-stereoset}. The NOI words are described in \Cref{sec:Appendix} (Table \ref{tab:identity_groups}).
\begin{table}[]
\centering
    \renewcommand{\arraystretch}{1.2}
    \resizebox{0.4\textwidth}{!}{
    \begin{tabular}{l|l|l} 
    \hline
         {Profane sentence} & Non-profane sentence & Identity\\ \hline
         \textcolor{red}{You are a dumb ....}	    & \textcolor{teal}{You are a friendly ....}  & Asian  \\ \hline
         \textcolor{red}{You are a moronic ....} & \textcolor{teal}{You are a incredible .... } & Asian\\ \hline
         \textcolor{red}{You are a fucked ....} &  \textcolor{teal}{You are a caring ....}  & Asian   \\ \hline
         \textcolor{red}{You are a awful ....} &	\textcolor{teal}{You are a best ....} & Asian\\ \hline

        \textcolor{red}{....} &	\textcolor{teal}{ ....} & .....\\ \hline
         \textcolor{red}{You are a dumb ....}	    & \textcolor{teal}{You are a friendly ....}  & Dutch  \\ \hline
         \textcolor{red}{You are a moronic ....} & \textcolor{teal}{You are a incredible .... } & Dutch\\ \hline
         \textcolor{red}{You are a fucked ....} &  \textcolor{teal}{You are a caring ....}  & Dutch   \\ \hline
         \textcolor{red}{You are a awful ....} &	\textcolor{teal}{You are a best ....} & Dutch\\ \hline
    \end{tabular}}
    \caption{\small{Examples of template profane/non-profane sentence-pairs. The identity column contains NOI words to describe both marginalized and non-marginalized identities to fill in the blanks in the profane and the non-profane sentences.}}
    \label{tab:list-of-sentence-pairs}
\end{table}

Then, to create a list of profane and non-profane sentence-pairs, we need a list of profane and non-profane words. We use GPT3 model, on the OpenAI platform\footnote{\url{https://platform.openai.com/overview}}, to automatically generate profane and non-profane words. We used the following prompts\footnote{We generated these words back in summer 2022. We acknowledge that the same prompt might generate different words.}: \textit{1) Write a list of offensive words} and \textit{2) Write a list of nice words}. GPT3 generated 21 profane and 21 non-profane words, which are then used with sentence templates to create the synthesized dataset.  Using the NOI words and the profane and non-profane word lists, we create a synthesized dataset that meets our criteria. Examples of the template sentence-pairs are provided in Table \ref{tab:list-of-sentence-pairs}. The final synthesized dataset contains 1638 sentence-pairs. 

\begin{figure*}[]
 \centering
    \includegraphics[width=0.6\textwidth]{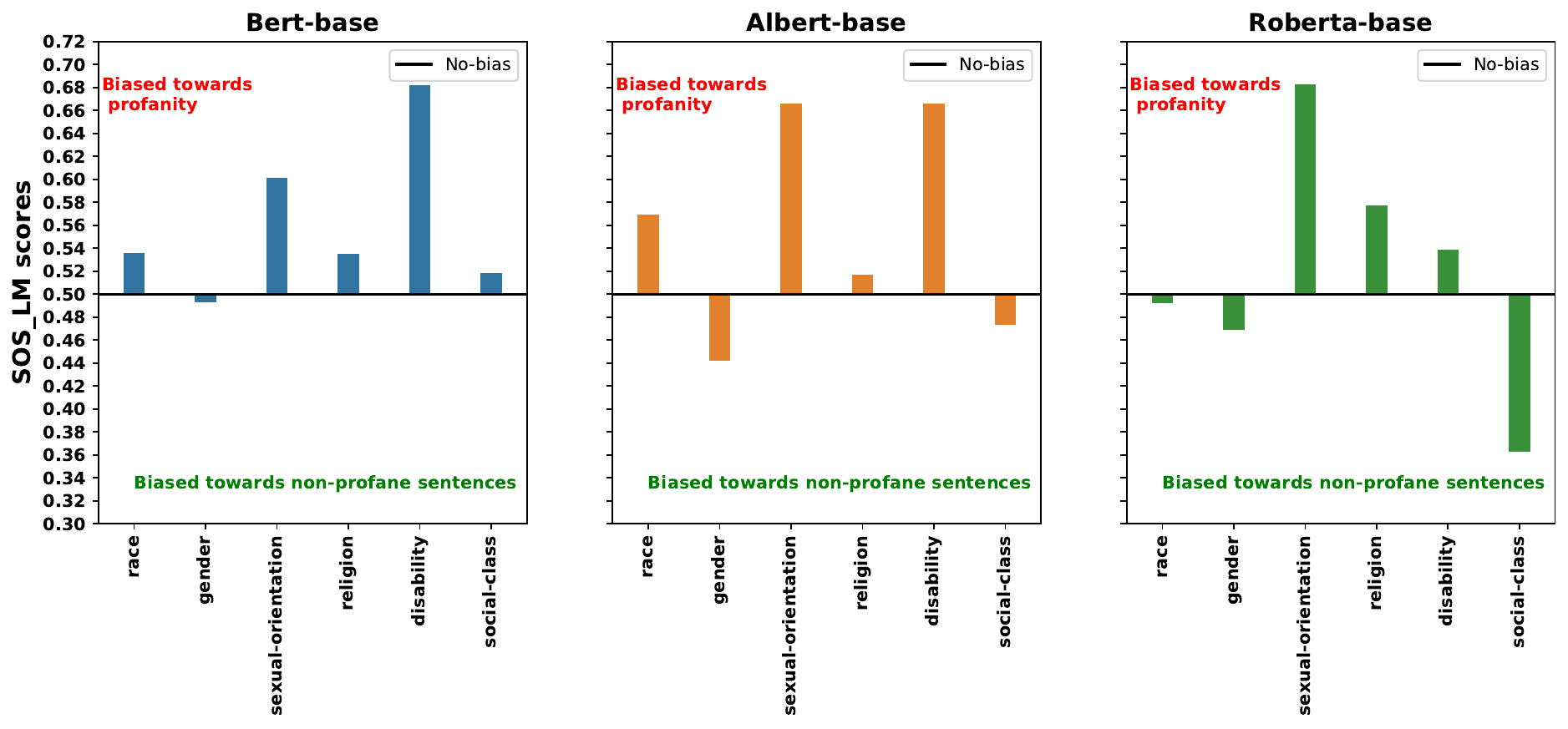}
    \caption{\small{$SOS_{LM}$ bias scores in the different LMs against all identity groups (marginalized and non-marginalized).}}
    \label{fig:SOS_bias_scores_all_attributes}
\end{figure*}
\begin{table*}[]
\centering
\renewcommand{\arraystretch}{1.2}
\resizebox{0.7\textwidth}{!}{
\centering
\begin{tabular}{l|lllllllllll}
\hline
             & \multicolumn{11}{c}{\textbf{SOS bias Scores}}                                                                                                                                                                                                                                                                                                                                                                  \\ \hline
Model        & \multicolumn{2}{l||}{Gender}                                               & \multicolumn{2}{l||}{Race}                                                 & \multicolumn{2}{l||}{Sexual-orientation}                                   & \multicolumn{2}{l||}{Religion}                                       & \multicolumn{2}{l||}{Social class}                                   & Disability \\ \hline
             & \multicolumn{1}{l|}{M}       & \multicolumn{1}{l||}{N}       & \multicolumn{1}{l|}{M}       & \multicolumn{1}{l||}{N}       & \multicolumn{1}{l|}{M}       & \multicolumn{1}{l||}{N}       & \multicolumn{1}{l|}{M} & \multicolumn{1}{l||}{N}       & \multicolumn{1}{l|}{M}       & \multicolumn{1}{l||}{N} & M   \\ \hline
BERT-base    & \multicolumn{1}{l|}{0.476}          & \multicolumn{1}{l||}{\textbf{0.510}} & \multicolumn{1}{l|}{\textbf{0.580}} & \multicolumn{1}{l||}{0.501}          & \multicolumn{1}{l|}{0.576}          & \multicolumn{1}{l||}{\textbf{0.714}} & \multicolumn{1}{l|}{0.523}    & \multicolumn{1}{l||}{\textbf{0.555}} & \multicolumn{1}{l|}{\textbf{0.560}} & \multicolumn{1}{l||}{0.480}    & \textbf{0.682}      \\ \hline
AlBERT-base  & \multicolumn{1}{l|}{\textbf{0.448}} & \multicolumn{1}{l||}{0.435}          & \multicolumn{1}{l|}{0.542}          & \multicolumn{1}{l||}{\textbf{0.589}} & \multicolumn{1}{l|}{\textbf{0.671}} & \multicolumn{1}{l||}{0.642}          & \multicolumn{1}{l|}{0.495}    & \multicolumn{1}{l||}{\textbf{0.555}} & \multicolumn{1}{l|}{\textbf{0.492}} & \multicolumn{1}{l||}{0.457}    & \textbf{0.666}      \\ \hline
RoBERTa-base & \multicolumn{1}{l|}{\textbf{0.517}} & \multicolumn{1}{l||}{0.421}          & \multicolumn{1}{l|}{\textbf{0.519}} & \multicolumn{1}{l||}{0.472}          & \multicolumn{1}{l|}{0.666}          & \multicolumn{1}{l||}{\textbf{0.761}} & \multicolumn{1}{l|}{0.561}    & \multicolumn{1}{l||}{\textbf{0.603}} & \multicolumn{1}{l|}{\textbf{0.391}} & \multicolumn{1}{l||}{0.338}    & \textbf{0.539}      \\ \hline
\end{tabular}}
\caption{\small{$SOS_{LM}$ scores of the different identity groups for all the language models. Bold values represent higher SOS bias scores between the marginalized (M) and the non-marginalized (N) groups in each sensitive attribute.}}
\label{tab:sos_bias_scors_comaprison_LM}
\end{table*}
\subsection{$SOS_{LM}$ bias metric}
\label{sec:sos-bias-metric}
To measure the SOS bias in LMs, we use the MLM task. For a profane sentence ($S$) where, $ S = U \cup M$, $U$ is a set of unmodified tokens for example,  $U = \{you, are, a, arab\}$ with length $|C|$, and $M$ is a set of modified tokens for example, $M = \{vile\}$. To estimate the probability of the unmodified token conditioned on the modified tokens $p(U| M, \theta)$, we use the \textit{pseudo-log-likelihood} \cite{nangia-etal-2020-crows}. The profane sentence $score(S)$ is then measured as:

 \begin{equation}
 \small
     score(S) = \sum_{i=0}^{|C|}{logP(u_i \in U | M, \theta)}
 \end{equation}

The same score is also measured for the non-profane sentence ($S'$) where $S' = U \cup M'$, $U$ is a set of unmodified tokens for example,  $U = \{you, are, a, arab\}$ with length $|C|$, and $M'$ is a set of modified tokens for example, $M' = \{nice\}$. 

  \begin{equation}
  \small
     score(S') = \sum_{i=0}^{|C|}{logP(u_i \in U | M', \theta)}
 \end{equation}

Then, the bias scores are measured as the percentage of examples where the model ($\theta$) assigns a higher probability estimate to the profane sentences ($S$) over the non-profane sentences ($S'$) as in equation \ref{eq:sos-bias-scores-lm} where ($N$) is the number of sentence-pairs. If the percentage is over or below 0.5, then that means the model prefers profane or non-profane sentences, respectively, and is biased. On the other hand, if the percentage is 0.5, that means the model randomly assigns probability and hence is not biased. Since the focus of this paper is to measure the offensive stereotyping bias, we only consider a LM to be SOS-biased, if the $SOS_{LM} > 0.5$.
 \begin{equation}
 \small
     SOS_{LM}= \frac{Count(score(S) > Score(S'))}{N}
     \label{eq:sos-bias-scores-lm}
 \end{equation}

\subsection{SOS biased LMs}
\label{sec:sos-bias-results}
We first measure the SOS bias scores against all the identity groups, marginalized and non-marginalized. The measured SOS bias scores in Figure \ref{fig:SOS_bias_scores_all_attributes} show that the majority of the inspected LMs are SOS biased, with ($SOS_{LM} > 0.5$), for the following sensitive attributes: race, sexual-orientation, religion, and disability. This indicates that the inspected LMs, in general, prefer profane sentences to non-profane ones. Then, we inspect the results closely to investigate whether the SOS bias scores in the inspected LMs are higher against the marginalized identity groups. We measure the SOS bias scores for the marginalized groups (M) and the non-marginalized groups (N) separately. Then, we compare the SOS bias scores between the marginalized and the non-marginalized identities. The results in Table \ref{tab:sos_bias_scors_comaprison_LM} show that the majority of the models have higher bias scores against the marginalized identity groups for the following sensitive attributes: gender, race, social class, and disability (not statistical significant difference at $\alpha = 0.05$).  We speculate that the higher SOS bias scores against marginalized groups could be a result of using biased pre-training datasets and an optimization method that might exacerbate that bias, as discussed by \citet{shah-etal-2020-predictive}. 

On the other hand, the majority of the models have higher SOS bias scores against the non-marginalized groups for the sexual-orientation and religion sensitive attributes (no statistical significant difference at $\alpha = 0.05$). We speculate that this is the case because LMs might be SOS biased against the attribute itself. In other words, LMs consider these topics to be taboos and associate profanity with any mention of any sexual orientation or religion, marginalized or not. 

\begin{table}[h]
\centering
    \renewcommand{\arraystretch}{1}
    \resizebox{0.3\textwidth}{!}{
\begin{tabular}{llll}
\hline
\multicolumn{4}{c}{\textbf{CrowS-Pairs}}                                                                                                                                                                                                                   \\ \hline
\multicolumn{1}{l|}{Bias}         & \multicolumn{1}{l|}{BERT}                                                 & \multicolumn{1}{l|}{RoBERTa}                                                  & \multicolumn{1}{l}{AlBERT}                            \\ \hline
    
\multicolumn{1}{l|}{Gender}   & \multicolumn{1}{l|}{{0.580}}          & \multicolumn{1}{l|}{\textbf{0.606}}             & \multicolumn{1}{l}{0.541}          \\ \hline
\multicolumn{1}{l|}{Race}     & \multicolumn{1}{l|}{\textbf{0.581}}           & \multicolumn{1}{l}{0.527}             & \multicolumn{1}{l}{0.513}         \\ \hline
\multicolumn{1}{l|}{Religion} & \multicolumn{1}{l|}{0.714}                    & \multicolumn{1}{l|}{\textbf{0.771}}             & \multicolumn{1}{l}{0.590}          \\ \hline

\multicolumn{4}{c}{\textbf{StereoSet}}                                                                                                                                                                                                                      \\ \hline
\multicolumn{1}{l|}{Bias}         & \multicolumn{1}{l|}{BERT}                                                 & \multicolumn{1}{l|}{RoBERTa}                                                  & \multicolumn{1}{l}{AlBERT}                            \\ \hline

\multicolumn{1}{l|}{Gender}   & \multicolumn{1}{l|}{0.602}          & \multicolumn{1}{l|}{\textbf{0.663}}            & \multicolumn{1}{l}{0.599}          \\ \hline
\multicolumn{1}{l|}{Race}     & \multicolumn{1}{l|}{0.570}          & \multicolumn{1}{l|}{\textbf{0.616}}            & \multicolumn{1}{l}{0.575}          \\ \hline
\multicolumn{1}{l|}{Religion} & \multicolumn{1}{l|}{0.597}           & \multicolumn{1}{l|}{\textbf{0.642}}             & \multicolumn{1}{l}{0.603}          \\ \hline

\multicolumn{4}{c}{\textbf{SEAT}}                                                                                                                                                                                                                          \\ \hline
\multicolumn{1}{l|}{Bias}         & \multicolumn{1}{l|}{BERT}                                                 & \multicolumn{1}{l|}{RoBERTa}                                                  & \multicolumn{1}{l}{AlBERT}                            \\ \hline

\multicolumn{1}{l|}{Gender}   & \multicolumn{1}{l|}{{0.620}}           & \multicolumn{1}{l|}{\textbf{0.939}}             & \multicolumn{1}{l}{{0.622}}              \\ \hline
\multicolumn{1}{l|}{Race}     & \multicolumn{1}{l|}{\textbf{0.620}}           & \multicolumn{1}{l|}{0.307}              & \multicolumn{1}{l}{{0.551}}             \\ \hline
\multicolumn{1}{l|}{Religion} & \multicolumn{1}{l|}{\textbf{0.491}}           & \multicolumn{1}{l|}{0.126}               & \multicolumn{1}{l}{0.430} \\ \hline

\end{tabular}}
\caption{\small Social bias scores in LMs. \textbf{Bold} scores mean higher bias scores and more biased models.}
\label{tab:socail_bias_scores}
\end{table}
\begin{figure}[t]
 \centering
    \includegraphics[width=0.5\textwidth]{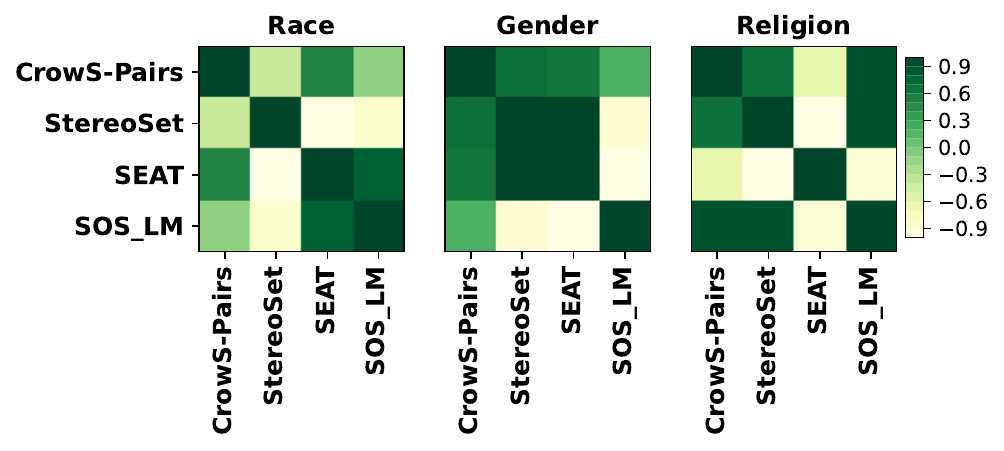}
    \caption{\small{Heatmap of the Pearson's correlation ($\rho$) between the $SOS_{LM}$ bias and social bias scores.}}
    \label{fig:heatmap_corr_sos_lm_social_bias} 
\end{figure}
\section{SOS bias validation}
We validate two aspects of the $SOS_{LM}$ bias metric. The first aspect is how different it is from social bias metrics proposed in the literature. The second aspect is how reflective it is of the online hate experienced by marginalized identity groups.

\subsection{SOS bias vs. social bias in LMs}
\label{sec:sos-bias-wv-social-bias}
We investigate the difference between the measured SOS bias scores and the social bias scores in the inspected LMs. We first measure the social bias scores in the LMs (Bert-base-uncased, AlBERT-base, and RoBERTa-base) using three bias metrics CrowS-Pairs \cite{nangia-etal-2020-crows}, StereoSet \cite{nadeem-etal-2021-stereoset}, and SEAT \cite{DBLP:conf/naacl/MayWBBR19}. The social bias scores are reported in Table \ref{tab:socail_bias_scores}.

To investigate the difference between social bias and SOS bias scores, we measure the Pearson correlation coefficient ($\rho$) between the SOS bias scores measured using the proposed $SOS_{LM}$ metric and the social bias scores measured using CrowS-Pairs, StereoSet, and SEAT metrics. The correlation is measured for three sensitive attributes: race, gender, and religion, as these attributes are common among all the used social bias metrics. Figure \ref{fig:heatmap_corr_sos_lm_social_bias} shows that, there is a positive correlation between the measured SOS bias scores and social bias scores measured using different bias metrics. However, the positive correlation is not consistent across the different sensitive attributes. 
The most consistent positive correlation is found between the SOS bias scores and the Crows-Pairs scores. This could be because our $SOS_{LM}$ metric uses a similar method to the CrowS-Pairs metric to measure SOS bias. These results suggest that, unlike the case with static word embeddings \cite{elsafoury_sos_2022}, our proposed metric to measure the SOS bias in LMs does not reveal different information from that revealed by social bias metrics, especially when measured using the CrowS-Pairs metric. 
  \begin{table}[h]
    \renewcommand{\arraystretch}{1}
    \centering
    \resizebox{0.9\columnwidth}{!}{
  \begin{tabular} {lcccc}
  \hline
   \textbf{Country} &
   \textbf{Sample size}&
   \textbf{Ethnicity} &
   \textbf{LGBTQ} &
   \textbf{Women} \\\hline
   Finland & 555	& 0.67	& 0.63	& 0.25 \\\hline
   US &	1033& 0.6	& 0.61	& 0.44 \\\hline
   Germany & 978	&0.48	& 0.5	& 0.2 \\\hline
   UK & 999	& 0.57	& 0.55	& 0.44 \\\hline
\end{tabular}}
\caption{\small{The percentage of examined marginalized groups that experience online hate and extremism \cite{hawdon2015online}}}
\label{tab:OHOE_data}
\end{table}
 \begin{figure}[h]
\centering
    \includegraphics[width=0.55
\columnwidth]{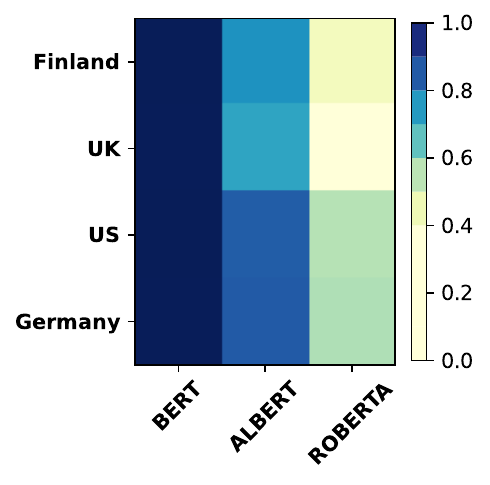}
    \caption{\small{Heat-map of the Pearson's correlation ($\rho$) between the SOS bias scores measured using the $SOS_{LM}$ metric and the percentages of marginalized identities who experience online hate in different countries.}
    \label{fig:heatmap_corr_sos_lm_OEOH}}
\end{figure}
\begin{table*}
\centering
    \renewcommand{\arraystretch}{1.1}
    \resizebox{0.75\textwidth}{!}{
\begin{tabular}{l|r|r|r|r|r|r|r|r|r|r|r|r}
\hline
Model & \multicolumn{3}{c|}{Crows-Pairs} & \multicolumn{3}{c|}{StereoSet} & \multicolumn{3}{c|} {SEAT}& \multicolumn{3}{c} {$SOS_{LM}$}\\ \hline

                              & Gender    & Race      & Religion  & Gender   & Race     & Religion   & Gender   & Race    & Religion & Gender   & Race    & Religion  \\\hline
\textbf{AlBERT-base}          & 0.541    & 0.513     & 0.590      & 0.599    & 0.575     & 0.603     & 0.622    & 0.551    & 0.430   &  0.440  & 0.570    & 0.520    \\ \hline

+ SentDebias-SOS         & \color{teal}$\downarrow0.503$    & \color{red}$\uparrow0.743$     & \color{red}$\uparrow0.714$      & \color{teal}$\downarrow0.504$    & \color{teal}$\downarrow0.468$     & \color{teal}$\downarrow0.539$     & 0.622    & 0.551    & 0.430   &  \color{red}$\uparrow0.639$  &  \color{teal}$\downarrow0.504$   &  \color{teal}$\downarrow0.485$    \\ \hline

\textbf{BERT-base-uncased}    & 0.580    & 0.581      & 0.714     & 0.607    & 0.570     & 0.597     & 0.620    & 0.620    & 0.491   & 0.490   &  0.540   & 0.540     \\ \hline

+ SentDebias-SOS         & \color{teal}$\downarrow0.572$    & \color{teal}$\downarrow0.473$     & \color{teal}$\downarrow0.609$      & \color{teal}$\downarrow0.485$    & \color{teal}$\downarrow0.430$     & \color{teal}$\downarrow0.436$     & 0.620    & 0.620    & 0.491   &  \color{red}$\uparrow0.782$  &  \color{red}$\uparrow0.581$   &  \color{teal}$\downarrow0.361$    \\ \hline

\textbf{RoBERTa-base}         & 0.606   & 0.527       & 0.771    & 0.663     & 0.616      & 0.642     & 0.939    & 0.307    & 0.126  & 0.470   &  0.490   &0.580        \\ \hline

+ SentDebias-SOS         & \color{teal}$\downarrow0.494$    & \color{red}$\uparrow0.567$     & \color{teal}$\downarrow0.361$      & \color{teal}$\downarrow0.517$    & \color{teal}$\downarrow0.463$     & \color{teal}$\downarrow0.457$     & 0.939    & 0.307    & 0.126   &  \color{red}$\uparrow0.734$  &  \color{teal}$\downarrow0.285$   &  \color{teal}$\downarrow0.438$    \\ \hline

\end{tabular}}
\caption{\small{Social and SOS bias scores in the different models using different bias metrics before and after removing SOS bias using SentDebias. (\textcolor{red}{$\uparrow$}) means that the bias score increased and the bias in the LMs worsened. (\textcolor{teal}{$\downarrow$}) means that the bias score decreased, and the bias in the LMs improved. The SOS bias scores reported here are against identity groups marginalized and non-marginalized.}}
\label{tab:intrinsic_bias_scores}
\end{table*}
\subsection{SOS bias and online hate}
\label{sec:sos-bias-vs-online-hate}
We investigate how reflective the SOS bias is, in the inspected LMs against marginalized identity groups, of the online hate that the same marginalized groups experience. We use published statistics of the percentages of marginalized groups that experience online hate and extremism \cite{hawdon2015online}. Table \ref{tab:OHOE_data} reports these statistics. We measure the Pearson correlation coefficients ($\rho$) between the online hate statistics and the SOS bias scores measured using our proposed $SOS_{LM}$ metric against the marginalized groups (M) in \Cref{tab:sos_bias_scors_comaprison_LM} for the following sensitive attributes: race, gender, and sexual-orientation. 

The results in Figure \ref{fig:heatmap_corr_sos_lm_OEOH}, show a strong positive correlation between the SOS bias measured in the inspected LMs using the proposed $SOS_{LM}$ metric and the published percentages of marginalized people who experience online hate and extremism in Finland, Germany, the US, and the UK. This strong positive correlation exists for BERT, followed by AlBERT and then RoBERTa. These results suggest that the proposed metric of measuring SOS bias in LMs is reflective of the hate that women, non-white ethnicities, and LGBTQ communities experience online. These results are inline with previous research on the SOS bias in static word embeddings \cite{elsafoury_sos_2022}.

\section{SOS bias removal}
\label{sec:sos-bias-removal}
We investigate the effectiveness of one of the state-of-the-art social bias removal methods in the literature, on removing the SOS bias in LMs. We use SentDebias \cite{liang2020towards} to remove different types of bias from the LMs by projecting a sentence representation onto the estimated bias subspace and subtracting the resulting projection from the original sentence representation. \citet{liang2020towards} compute the bias subspace by following these steps: 1) Define a list of identity words, e.g., \say{woman/man}; 2) Contextualize the identity words into sentences by finding sentences that contain those identity words in public datasets like SST\footnote{\url{https://huggingface.co/datasets/sst}} and WikiText-2\footnote{\url{https://huggingface.co/datasets/wikitext}}; 3) Obtain the representation of the contextualized sentence from the pre-trained LM; 4) Principal Component Analysis (PCA) \cite{abdi2010principal} then used to estimate principal directions of variations of the sentences' representations. The first $K$ principal components are taken to define the bias subspace.  For removing the SOS bias, we follow the same debias steps as \cite{liang2020towards}. But instead of computing the social bias subspace, we compute the profanity subspace. We first contextualize the 21 profane and 21 non-profane words used in section \ref{sec:sos_bias_dataset} using the WikiText-2 dataset as explained earlier. Then we remove the profanity subspace from the inspected LMs. We build on the implementation shared by \cite{meade-etal-2022-emprical-survey-debias} to remove gender, racial, religion, and SOS bias.

The results, in \Cref{tab:intrinsic_bias_scores}, 
show that, in some cases, removing the SOS bias improved the social bias scores according to CrowS-Pairs and StereoSet. On the other hand, according $SOS_{LM}$, SentDebias improved the SOS bias scores for some sensitive attributes (race and religion) and worsened the bias in other sensitives attributes (gender). We found the same results for the SOS bias scores when measured against all identity groups
, SOS bias scores measured against marginalized identities (M) and SOS bias scores measured against non-marginalized identities.
The SEAT metric, did not show any difference in the bias scores for the debiased models, unlike the reported scores in \cite{meade-etal-2022-emprical-survey-debias}. 

Then, we calculate the T-test statistical significance test between the two independent samples of bias scores before and after applying the SentDebias algorithm to remove the SOS bias. We use the bias scores as measured by the CrowS-Pairs, StereoSet and $SOS_{LM}$ metrics since these are the metrics that have different results after removing the SOS bias. The results show that according to StereoSet removing the SOS bias significantly improved the social bias scores for AlBERT ($pvalues = 0.01$), BERT ($p-value = 0.002$), and RoBERTa ($pvalue = 0.002$) at $\alpha = 0.05$.  

So far, we introduced the $SOS_{LM}$ metric to measure the SOS bias in LMs, validated it, and investigated the effectiveness of its removal. In the rest of this paper, we investigate the impact of the SOS bias in LMs on the performance and fairness of the downstream task of hate speech detection.

\section{Impact of SOS bias on the performance of hate speech detection}
\label{sec:sos-impact-on-performance}
 To evaluate the performance of the inspected LMs on the task of hate speech detection, we first fine-tune the inspected LMs on the hate speech related datasets described in \Cref{tab:data-sats-ch5-lm}.

We follow the same pre-processing steps described in \cite{elsafoury2021}, as the authors fine-tune BERT on the task of hate speech detection, which are:
(1) remove URLs, user mentions, non-ASCII characters, and the retweet abbreviation \say{RT} (Twitter datasets). (2) All letters are lower cased. (3) Contractions are converted to their formal format. (4) A space is added between words and punctuation marks. We then split the datasets into 40\% training set, 30\% validation set and 30\% test set. We train the models for 3 epochs, using a batch size of 32, a learning rate of $2e^{-5}$, and a maximum text length of 61 tokens. The performance results (F1-scores) are reported in Table \ref{tab:f1_scores_LM}.
\begin{table}[]
\centering
    \renewcommand{\arraystretch}{1.2}
    \resizebox{0.85\columnwidth}{!}{
  \begin{tabular} {l|c|c|c}
  \hline
  \multirow{2}{*}{\textbf{Dataset}} &
    \multirow{2}{*}{\textbf{Samples}} &
    \textbf{Positive} \\
    & & \textbf{samples} &
     \\ \hline
    Twitter-sexism & 14742 &23\% & \cite{waseem_hateful_2016} \\\hline
    Twitter-racism & 13349 & 15\%  & \cite{waseem_hateful_2016} \\\hline
    Civil-community & 426707 & 0.08\%  & \cite{elsafoury2023bias} \\ \hline
    Kaggle-insults & 7425 & 35\% & \cite{kaggle-insults} \\ \hline
    WTP-agg & 114649 & 13\% & \cite{wulczyn_ex_2017}\\ \hline
    WTP-tox & 157671 & 10\% &  \cite{wulczyn_ex_2017}\\ \hline
   \multicolumn{3}{l}{\footnotesize Note: Positive samples refer to offensive comments}
\end{tabular}}
\caption{\small{Statistics of hate speech datasets used with the inspected language models.}}
 \label{tab:data-sats-ch5-lm}
\end{table}
\begin{table}[]
    \centering
        \renewcommand{\arraystretch}{1}
    \resizebox{0.75\columnwidth}{!}{
    \begin{tabular}{c|c|c|c} \hline
    Dataset & BERT & AlBERT& ROBERTA  \\ \hline
    Kaggle & 0.844 & 0.832	& \textbf{0.847} \\ \hline
    Twitter-sexism &	0.871&	\textbf{0.884} &	0.880 \\ \hline
    Twitter-racism	& \textbf{0.930} &	0.924 &	0.929 \\ \hline
    WTP-agg	& 0.937 &	\textbf{0.939} &	0.934 \\ \hline
    WTP-tox & 0.960	& 0.961 &	\textbf{0.963} \\ \hline
    Civil-community&	0.582	& 0.558	& \textbf{0.589} \\ \hline
    \end{tabular}}
    \caption{\small{F1 scores of the inspected LMs on the different hate speech datasets. \textbf{Bold} values denote the best performance.}}
    \label{tab:f1_scores_LM}
\end{table}
Then, to investigate the impact of the SOS bias in the LMs on their performance on hate speech detection, we use correlation. We compute the Pearson correlation coefficient ($\rho$) between the F1-scores of the  LMs reported in Table \ref{tab:f1_scores_LM} and the SOS bias scores against the marginalized identities (M) displayed in Table \ref{tab:sos_bias_scors_comaprison_LM}. The results, in Table \ref{tab:perason_corr_metrics_f1_scores_lm}, show a strong positive correlation with the SOS bias scores against marginalized identities in all the datasets: Twitter-racism (race, gender, and religion); WTP-agg (race, disability, and social class); and WTP-toxicity (gender, sexual-orientation, and religion); Kaggle (gender, religion); Civil-community (gender, and religion); and Twitter-sexism (sexual-orientation). However, these results are not consistent across all the sensitive attributes. We speculate that this is due to the different targets of the hate in the different hate speech datasets. For example, the SOS bias scores in the gender, race, sexual-orientation, and religion sensitive attributes, correlate positively with the performance of the LMs on the following hate speech dataset: Twitter-racism, WTP-aggression, and WTP-toxicity where the targets of the hate in the datasets are matching the marginalized groups in the gender, race, sexual-orientation, and religion sensitive attributes. Yet, these results and the impact of the SOS bias on the performance of hate speech detection remain inconclusive. 

\begin{table}[]
    \renewcommand{\arraystretch}{1.3}
    \resizebox{0.47\textwidth}{!}{
\begin{tabular}{l|r|r|r|r|r|r}
\hline
{} &  \multicolumn{5}{c}{Sensitive attribute} \\
\hline
Dataset        & \multicolumn{1}{l|}{Race} & \multicolumn{1}{l|}{Gender} & \multicolumn{1}{l|}{Sexuality} & \multicolumn{1}{l|}{Religion} & \multicolumn{1}{l|}{Disability} & \multicolumn{1}{l}{Social class} \\ \hline
Kaggle         & -0.049                    & {0.903}              & -0.371                                  & {0.912}                & -0.574                          & -0.297                            \\ \hline
Twitter-sexism & -0.772                    & -0.195                      & {0.966}                          & -0.216                        & -0.315                          & -0.589                            \\ \hline
Twitter-racism & {0.292}            & {0.705}              & -0.664                                  & {0.719}                & -0.262                          & 0.043                             \\ \hline
WTP-agg        & {0.477}            & -0.999                      & -0.068                                  & -0.999                        & {0.872}                  & {0.682}                    \\ \hline
WTP-toxicity   & -0.945                    & {0.732}              & {0.724}                          & {0.718}                & -0.973                          & -0.996                            \\ \hline
Civil-community     & -0.075                    & {0.915}              & -0.346                                  & {0.923}                & -0.595                          & -0.323                            \\ \hline
\end{tabular}}
\caption{\small{Pearson Correlation Coefficient ($\rho$) between the SOS bias scores against the marginalized groups in the inspected LMs and the F1 scores of the different LMs on each dataset.}}
\label{tab:perason_corr_metrics_f1_scores_lm}
\end{table}
\section{Impact of SOS bias on fairness of hate speech detection}
\label{sec:sos-bias-impact-on-fairness}
To measure the impact of the SOS bias in the inspected LMs on their fairness of the task of hate speech detection, we first need to measure the fairness of the inspected LMs on hate speech detection. To this end, we use the fairness scores reported in \cite{elsafoury2023bias}.  Where the authors measure fairness as the absolute difference in the false positive rates ($FPR$), true positive rates ($TPR$), and the area under the curve ($AUC$) between the marginalized group ($g$) and non-marginalized group ($\hat{g}$), as shown in \cref{eq:fpr-gap}, \cref{eq:tpr-gap}, and \cref{eq:auc-gap}.  These scores measure the unfairness of the model in how it treats different identity groups of people differently. The higher the score, the more unfair the model is and the lower the scores, the better the model in terms of fairness. Table \ref{tab:senstive_attributs} describes the inspected identity groups in three sensitive attributes: gender, race, and religion. \citet{elsafoury2023bias} measure fairness of the same LMs that we use in this work, AlBERT-base, BERT-base, and RoBERTa-base, on the downstream task of hate speech detection using the Civil comments dataset described in Table \ref{tab:data-sats-ch5-lm}.

\begin{equation}
\small
    FPR\_gap_{g, \hat{g}} = |FPR_{g} - FPR_{\hat{g}}|  
    \label{eq:fpr-gap}
\end{equation}
\begin{equation}
\small
    TPR\_gap_{g, \hat{g}} = |TPR_{g} - TPR_{\hat{g}}|  
    \label{eq:tpr-gap}
\end{equation}
\begin{equation}
\small
    AUC\_gap_{g, \hat{g}} = |AUC_{g} - AUC_{\hat{g}}|  
    \label{eq:auc-gap}
\end{equation}
\begin{table}[h]
\centering
             \renewcommand{\arraystretch}{1}
    \resizebox{0.8\columnwidth}{!}{
     \begin{tabular}{l|l|l} \hline
         Sensitive attribute & Marginalized & Non-marginalized \\ \hline
         Gender & Female & Male \\ \hline
         Race & Black and Asian & White \\ \hline
         Religion &Jewish and Muslim & Christian\\ \hline
     \end{tabular}}
     \caption{{The inspected identity groups to measure fairness.}}
     \label{tab:senstive_attributs}
 \end{table}
\begin{table}[h]
    \centering
    \renewcommand{\arraystretch}{1}
    \resizebox{0.4\textwidth}{!}{
    \begin{tabular}{llrrr}
    \hline
         Attribute & Model   & FPR\_gap & TPR\_gap & AUC\_gap  \\
    \hline
         \multirow{5}{*}{Gender} &
         \multirow{1}{*}{AlBERT} &
       
     0.006    &  0.038   & \color{teal}{0.003} \\
    \cline{2-5}
    &     \multirow{1}{*}{BERT}

    &    0.008    &  0.036   &  0.009 \\
    \cline{2-5}
    &     \multirow{1}{*}{RoBERTa}

    &   \color{teal}{0.004}    &  \color{teal}{0.031}   &  0.011 \\
    \hline
         \multirow{5}{*}{Race} &
         \multirow{1}{*}{AlBERT} &

      0.008   &   \color{teal}0.001   &  \color{teal}{0.018} \\
    \cline{2-5}
    &     \multirow{1}{*}{BERT}

    &    0.015   &   {0.002}   &  0.025 \\
    \cline{2-5}
    &     \multirow{1}{*}{RoBERTa}

    &  \color{teal}{0.003}   &   0.011  &  0.021 \\
    \hline
         \multirow{5}{*}{Religion} &
         \multirow{1}{*}{AlBERT} &

       0.009  &   0.108   &  0.020 \\
    \cline{2-5}
    &     \multirow{1}{*}{BERT}

    &    \color{teal}{0.008}  &   \color{teal}{0.062}   &  \color{teal}{0.012} \\
    \cline{2-5}
    &     \multirow{1}{*}{RoBERTa}

    &     0.021  &   0.160   &  0.027 \\
    \hline
    \end{tabular}
    } 
    \caption{\small The fairness scores of the inspected LMs on the task of hate speech detection. \color{teal}{Teal color} \color{black} denotes the most fair model for each sensitive attribute according to each fairness metric \cite{elsafoury2023bias}. }
    \label{tab:extrinsic_bias_scores_perturbed_vs_original_jigsaw_dataset}
\end{table}
The fairness scores reported in Table \ref{tab:extrinsic_bias_scores_perturbed_vs_original_jigsaw_dataset} show that different fairness metrics give different fairness scores. However, there is a general trend that for the gender sensitive attribute, RoBERTa is the fairest, as for the race sensitive attribute, AlBERT is the fairest, and for the religion sensitive attribute, BERT is the fairest  according to the majority of the fairness metrics.

\begin{figure}
    \centering
    \includegraphics[width=0.6\columnwidth]{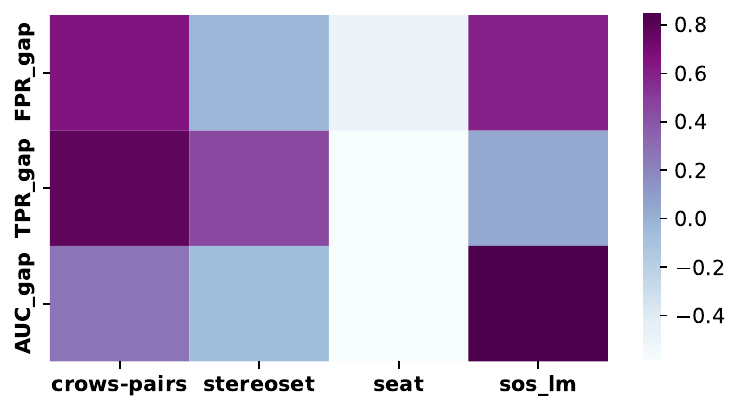}
    \caption{\small{Heatmap of Pearson's correlation between social and SOS bias in LMs and fairness scores of LMs on the downstream task of hate speech detection.}}
    \label{fig:intrinsic_vs_extrinsic_bias}
\end{figure}

To measure the impact of the SOS bias on the fairness scores, we measure the Pearson's correlation coefficient ($\rho$) between fairness scores measured by the different fairness metrics and the $SOS_{LM}$ bias scores. Additionally, we measure the impact of the social bias in the LMs as measured by the CrowS-Pairs, StereoSet, and SEAT metric and the fairness of the LMs on the task of hate speech detection.  The correlation results in \Cref{fig:intrinsic_vs_extrinsic_bias} show a consistent strong positive correlation between the CrowS-Pairs bias scores with the fairness scores measured by all three fairness metrics (FPR\_gap. TPR\_gap and AUC\_gap) for all the models and sensitive attributes. And a less strong positive correlation with TPR\_gap. There is a consistent negative correlation between SEAT scores and all fairness metrics. On the other hand, there is an inconsistent correlation with the StereoSet scores. As for the SOS bias, we find a positive correlation between the $SOS_{LM}$ bias scores, against marginalized identities, and the fairness scores as measured by the FPR\_gap and the AUC\_gap. The results of this section suggest that the SOS bias in the LMs as measured by $SOS_{LM}$ has an impact on the fairness of the LMs on the task of hate speech detection.

\section{Limitations}
One of the main limitations of this work is that we only study bias in Western societies where Women, LGBTQ and Non-White ethnicities are among the marginalized groups. However, marginalized groups could include different groups of people in other societies. We also only use datasets, word lists, and LMs in English, which limits our study to the English-speaking world. Another limitation is that the correlation scores reported in the paper are not statistically significant, which could be due to having a few data points. Another limitation of this work is using a generative model to generate a list of profane and non-profane words. These generated word lists might be biased and might have an impact on the measured SOS bias scores. Moreover, the use of template sentence-pairs to measure the SOS bias in LMs do not provide a real context which might have impacted the measured SOS bias. The findings of this work are limited to the examined word embeddings, models, datasets, word lists, and might not generalize to others.


\section{Conclusion}
In this paper, we build on existing social bias metrics, and propose the $SOS_{LM}$ metric to measure the systematic offensive stereotyping (SOS) bias in Language models (LMs) regarding six different sensitive attributes.  Our results show that all the inspected LMs are SOS biased and that for the majority of the sensitive attributes, the SOS bias in the LMs is higher against the marginalized groups. Then, we validate the proposed $SOS_{LM}$ metric in comparison to social bias metrics and published statistics on online hate that marginalized groups experience. Our results show that the proposed $SOS_{LM}$ metric does not reveal different information from the social bias metric, especially CrowS-Pairs. But our results also show that the proposed metric to measure the SOS bias in LMs is reflective of the online hate experienced by marginalized groups online. Subsequently, we use a state-of-the-art debias method, SentDebias, to remove the SOS bias. However, we found that SentDebias improved the SOS bias scores for some sensitive attributes and improved it for others. Finally, we investigate the impact of the SOS bias in LMs on their performance and fairness on hate speech detection. Our results suggest that there is an impact of the SOS bias in the LMs on their fairness on hate speech detection. However, there is no strong evidence that SOS bias has an impact on the performance of hate speech detection.

\bibliography{custom}

\begin{thebibliography}{28}
\expandafter\ifx\csname natexlab\endcsname\relax\def\natexlab#1{#1}\fi

\bibitem[{Abdi and Williams(2010)}]{abdi2010principal}
Herv{\'e} Abdi and Lynne~J Williams. 2010.
\newblock Principal component analysis.(2010).
\newblock \emph{Computational Statistics, John Wiley and Sons}, pages 433--459.

\bibitem[{Antoniak and Mimno(2021)}]{antoniak2021}
Maria Antoniak and David Mimno. 2021.
\newblock \href {https://doi.org/10.18653/v1/2021.acl-long.148} {Bad seeds: Evaluating lexical methods for bias measurement}.
\newblock In \emph{Proceedings of the 59th Annual Meeting of the Association for Computational Linguistics and the 11th International Joint Conference on Natural Language Processing (Volume 1: Long Papers)}, pages 1889--1904, Online. Association for Computational Linguistics.

\bibitem[{Borkan et~al.(2019)Borkan, Dixon, Sorensen, Thain, and Vasserman}]{Borkan-etal-2019-naunced-metrics}
Daniel Borkan, Lucas Dixon, Jeffrey Sorensen, Nithum Thain, and Lucy Vasserman. 2019.
\newblock \href {https://doi.org/10.1145/3308560.3317593} {Nuanced metrics for measuring unintended bias with real data for text classification}.
\newblock In \emph{WWW '19: Companion Proceedings of The 2019 World Wide Web Conference}, pages 491--500.

\bibitem[{Caliskan et~al.(2017)Caliskan, Bryson, and Narayanan}]{Caliskan2017}
Aylin Caliskan, Joanna~J. Bryson, and Arvind Narayanan. 2017.
\newblock \href {https://doi.org/10.1126/science.aal4230} {Semantics derived automatically from language corpora contain human-like biases}.
\newblock \emph{Science}, 356(6334):183--186.

\bibitem[{Devlin et~al.(2019)Devlin, Chang, Lee, and Toutanova}]{DBLP:conf/naacl/DevlinCLT19}
Jacob Devlin, Ming{-}Wei Chang, Kenton Lee, and Kristina Toutanova. 2019.
\newblock {BERT:} pre-training of deep bidirectional transformers for language understanding.
\newblock In \emph{{NAACL-HLT} {(1)}}, pages 4171--4186. Association for Computational Linguistics.

\bibitem[{Dixon et~al.(2018)Dixon, Li, Sorensen, Thain, and Vasserman}]{Dixon2018}
Lucas Dixon, John Li, Jeffrey Sorensen, Nithum Thain, and Lucy Vasserman. 2018.
\newblock \href {https://doi.org/10.1145/3278721.3278729} {Measuring and mitigating unintended bias in text classification}.
\newblock In \emph{Proceedings of the 2018 AAAI/ACM Conference on AI, Ethics, and Society}, AIES '18, page 67–73, New York, NY, USA. Association for Computing Machinery.

\bibitem[{Elsafoury(2022)}]{DBLP:conf/acl/Elsafoury22}
Fatma Elsafoury. 2022.
\newblock \href {https://doi.org/10.18653/v1/2022.acl-srw.4} {Darkness can not drive out darkness: Investigating bias in hate speechdetection models}.
\newblock In \emph{Proceedings of the 60th Annual Meeting of the Association for Computational Linguistics: Student Research Workshop, {ACL} 2022, Dublin, Ireland, May 22-27, 2022}, pages 31--43. Association for Computational Linguistics.

\bibitem[{Elsafoury et~al.(2023)Elsafoury, Katsigiannis, and Ramzan}]{elsafoury2023bias}
Fatma Elsafoury, Stamos Katsigiannis, and Naeem Ramzan. 2023.
\newblock On bias and fairness in nlp: How to have a fairer text classification?
\newblock \emph{arXiv preprint arXiv:2305.12829}.

\bibitem[{Elsafoury et~al.(2021)Elsafoury, Katsigiannis, Wilson, and Ramzan}]{elsafoury2021}
Fatma Elsafoury, Stamos Katsigiannis, Steven~R. Wilson, and Naeem Ramzan. 2021.
\newblock \href {https://doi.org/10.1145/3404835.3463029} {Does {BERT} pay attention to cyberbullying?}
\newblock In \emph{Proceedings of the 44th International ACM SIGIR Conference on Research and Development in Information Retrieval}, page 1900–1904, New York, NY, USA. Association for Computing Machinery.

\bibitem[{Elsafoury et~al.(2022)Elsafoury, Wilson, Katsigiannis, and Ramzan}]{elsafoury_sos_2022}
Fatma Elsafoury, Steve~R. Wilson, Stamos Katsigiannis, and Naeem Ramzan. 2022.
\newblock \href {https://aclanthology.org/2022.coling-1.108} {{SOS}: Systematic offensive stereotyping bias in word embeddings}.
\newblock In \emph{Proceedings of the 29th International Conference on Computational Linguistics}, pages 1263--1274, Gyeongju, Republic of Korea. International Committee on Computational Linguistics.

\bibitem[{Goldfarb-Tarrant et~al.(2021)Goldfarb-Tarrant, Marchant, Mu{\~n}oz~S{\'a}nchez, Pandya, and Lopez}]{goldfarb-tarrant-etal-2021-intrinsic}
Seraphina Goldfarb-Tarrant, Rebecca Marchant, Ricardo Mu{\~n}oz~S{\'a}nchez, Mugdha Pandya, and Adam Lopez. 2021.
\newblock \href {https://doi.org/10.18653/v1/2021.acl-long.150} {Intrinsic bias metrics do not correlate with application bias}.
\newblock In \emph{Proceedings of the 59th Annual Meeting of the Association for Computational Linguistics and the 11th International Joint Conference on Natural Language Processing (Volume 1: Long Papers)}, pages 1926--1940, Online. Association for Computational Linguistics.

\bibitem[{Hawdon et~al.(2015)Hawdon, Oksanen, and R{\"a}s{\"a}nen}]{hawdon2015online}
James Hawdon, Atte Oksanen, and Pekka R{\"a}s{\"a}nen. 2015.
\newblock Online extremism and online hate.
\newblock \emph{Nordicom-Information}, 37:29--37.

\bibitem[{Kaggle(2012)}]{kaggle-insults}
Kaggle. 2012.
\newblock Detecting insults in social commentary.
\newblock \url{https://www.kaggle.com/c/detecting-insults-in-social-commentary/data}.
\newblock {A}ccessed: 2020-09-28.

\bibitem[{Lan et~al.(2020)Lan, Chen, Goodman, Gimpel, Sharma, and Soricut}]{albert}
Zhenzhong Lan, Mingda Chen, Sebastian Goodman, Kevin Gimpel, Piyush Sharma, and Radu Soricut. 2020.
\newblock \href {https://openreview.net/forum?id=H1eA7AEtvS} {{ALBERT:} {A} lite {BERT} for self-supervised learning of language representations}.
\newblock In \emph{8th International Conference on Learning Representations, {ICLR} 2020, Addis Ababa, Ethiopia, April 26-30, 2020}. OpenReview.net.

\bibitem[{Liang et~al.(2020)Liang, Li, Zheng, Lim, Salakhutdinov, and Morency}]{liang2020towards}
Paul~Pu Liang, Irene~Mengze Li, Emily Zheng, Yao~Chong Lim, Ruslan Salakhutdinov, and Louis-Philippe Morency. 2020.
\newblock \href {https://doi.org/10.18653/v1/2020.acl-main.488} {Towards debiasing sentence representations}.
\newblock In \emph{Proceedings of the 58th Annual Meeting of the Association for Computational Linguistics}, pages 5502--5515, Online. Association for Computational Linguistics.

\bibitem[{Liu et~al.(2019)Liu, Ott, Goyal, Du, Joshi, Chen, Levy, Lewis, Zettlemoyer, and Stoyanov}]{roberta}
Yinhan Liu, Myle Ott, Naman Goyal, Jingfei Du, Mandar Joshi, Danqi Chen, Omer Levy, Mike Lewis, Luke Zettlemoyer, and Veselin Stoyanov. 2019.
\newblock \href {http://arxiv.org/abs/1907.11692} {Roberta: {A} robustly optimized {BERT} pretraining approach}.
\newblock \emph{CoRR}, abs/1907.11692.

\bibitem[{May et~al.(2019)May, Wang, Bordia, Bowman, and Rudinger}]{DBLP:conf/naacl/MayWBBR19}
Chandler May, Alex Wang, Shikha Bordia, Samuel~R. Bowman, and Rachel Rudinger. 2019.
\newblock \href {https://doi.org/10.18653/v1/n19-1063} {On measuring social biases in sentence encoders}.
\newblock In \emph{Proceedings of the 2019 Conference of the North American Chapter of the Association for Computational Linguistics: Human Language Technologies, {NAACL-HLT} 2019, Minneapolis, MN, USA, June 2-7, 2019, Volume 1 (Long and Short Papers)}, pages 622--628. Association for Computational Linguistics.

\bibitem[{Meade et~al.(2022)Meade, Poole-Dayan, and Reddy}]{meade-etal-2022-emprical-survey-debias}
Nicholas Meade, Elinor Poole-Dayan, and Siva Reddy. 2022.
\newblock \href {https://doi.org/10.18653/v1/2022.acl-long.132} {An empirical survey of the effectiveness of debiasing techniques for pre-trained language models}.
\newblock In \emph{Proceedings of the 60th Annual Meeting of the Association for Computational Linguistics (Volume 1: Long Papers)}, pages 1878--1898, Dublin, Ireland. Association for Computational Linguistics.

\bibitem[{Nadeem et~al.(2021)Nadeem, Bethke, and Reddy}]{nadeem-etal-2021-stereoset}
Moin Nadeem, Anna Bethke, and Siva Reddy. 2021.
\newblock \href {https://doi.org/10.18653/v1/2021.acl-long.416} {{S}tereo{S}et: Measuring stereotypical bias in pretrained language models}.
\newblock In \emph{Proceedings of the 59th Annual Meeting of the Association for Computational Linguistics and the 11th International Joint Conference on Natural Language Processing (Volume 1: Long Papers)}, pages 5356--5371, Online. Association for Computational Linguistics.

\bibitem[{Nangia et~al.(2020)Nangia, Vania, Bhalerao, and Bowman}]{nangia-etal-2020-crows}
Nikita Nangia, Clara Vania, Rasika Bhalerao, and Samuel~R. Bowman. 2020.
\newblock \href {https://doi.org/10.18653/v1/2020.emnlp-main.154} {{C}row{S}-pairs: A challenge dataset for measuring social biases in masked language models}.
\newblock In \emph{Proceedings of the 2020 Conference on Empirical Methods in Natural Language Processing (EMNLP)}, pages 1953--1967, Online. Association for Computational Linguistics.

\bibitem[{Nozza et~al.(2021)Nozza, Bianchi, and Hovy}]{nozza-etal-2021-honest}
Debora Nozza, Federico Bianchi, and Dirk Hovy. 2021.
\newblock \href {https://doi.org/10.18653/v1/2021.naacl-main.191} {{HONEST}: Measuring hurtful sentence completion in language models}.
\newblock In \emph{Proceedings of the 2021 Conference of the North American Chapter of the Association for Computational Linguistics: Human Language Technologies}, pages 2398--2406, Online. Association for Computational Linguistics.

\bibitem[{Nozza et~al.(2022)Nozza, Bianchi, Lauscher, and Hovy}]{nozza-etal-2022-measuring}
Debora Nozza, Federico Bianchi, Anne Lauscher, and Dirk Hovy. 2022.
\newblock \href {https://doi.org/10.18653/v1/2022.ltedi-1.4} {Measuring harmful sentence completion in language models for {LGBTQIA}+ individuals}.
\newblock In \emph{Proceedings of the Second Workshop on Language Technology for Equality, Diversity and Inclusion}, pages 26--34, Dublin, Ireland. Association for Computational Linguistics.

\bibitem[{Ousidhoum et~al.(2021)Ousidhoum, Zhao, Fang, Song, and Yeung}]{ousidhoum-etal-2021-probing}
Nedjma Ousidhoum, Xinran Zhao, Tianqing Fang, Yangqiu Song, and Dit-Yan Yeung. 2021.
\newblock \href {https://doi.org/10.18653/v1/2021.acl-long.329} {Probing toxic content in large pre-trained language models}.
\newblock In \emph{Proceedings of the 59th Annual Meeting of the Association for Computational Linguistics and the 11th International Joint Conference on Natural Language Processing (Volume 1: Long Papers)}, pages 4262--4274, Online. Association for Computational Linguistics.

\bibitem[{Shah et~al.(2020)Shah, Schwartz, and Hovy}]{shah-etal-2020-predictive}
Deven~Santosh Shah, H.~Andrew Schwartz, and Dirk Hovy. 2020.
\newblock \href {https://doi.org/10.18653/v1/2020.acl-main.468} {Predictive biases in natural language processing models: A conceptual framework and overview}.
\newblock In \emph{Proceedings of the 58th Annual Meeting of the Association for Computational Linguistics}, pages 5248--5264, Online. Association for Computational Linguistics.

\bibitem[{Sweeney and Najafian(2019)}]{sweeney2019}
Chris Sweeney and Maryam Najafian. 2019.
\newblock \href {https://doi.org/10.18653/v1/P19-1162} {A transparent framework for evaluating unintended demographic bias in word embeddings}.
\newblock In \emph{Proceedings of the 57th Annual Meeting of the Association for Computational Linguistics}, pages 1662--1667, Florence, Italy. Association for Computational Linguistics.

\bibitem[{Waseem and Hovy(2016)}]{waseem_hateful_2016}
Zeerak Waseem and Dirk Hovy. 2016.
\newblock \href {https://doi.org/10.18653/v1/n16-2013} {Hateful symbols or hateful people? predictive features for hate speech detection on twitter}.
\newblock In \emph{Proceedings of the Student Research Workshop, SRW@HLT-NAACL 2016, The 2016 Conference of the North American Chapter of the Association for Computational Linguistics: Human Language Technologies, San Diego California, USA, June 12-17, 2016}, pages 88--93. The Association for Computational Linguistics.

\bibitem[{Wulczyn et~al.(2017)Wulczyn, Thain, and Dixon}]{wulczyn_ex_2017}
Ellery Wulczyn, Nithum Thain, and Lucas Dixon. 2017.
\newblock Ex machina: Personal attacks seen at scale.
\newblock In \emph{Proceedings of the 26th International Conference on World Wide Web}, WWW '17, page 1391–1399. International World Wide Web Conferences Steering Committee.

\bibitem[{Zhu et~al.(2023)Zhu, Yuan, Wang, Liu, Liu, Deng, Dou, and rong Wen}]{Zhu2023LargeLM}
Yutao Zhu, Huaying Yuan, Shuting Wang, Jiongnan Liu, Wenhan Liu, Chenlong Deng, Zhicheng Dou, and Ji~rong Wen. 2023.
\newblock \href {https://api.semanticscholar.org/CorpusID:260887838} {Large language models for information retrieval: A survey}.
\newblock \emph{ArXiv}, abs/2308.07107.

\end{thebibliography}
\bibliographystyle{acl_natbib}

\appendix
\section{Appendix}
\label{sec:Appendix}
\begin{table*}[]
\centering
    \renewcommand{\arraystretch}{1.2}
     \resizebox{0.8\textwidth}{!}{
\begin{tabular}{l|l|l}
\hline
Attribute & Marginalized                                                                                                                                   & Non-marginalized                                                                                                                                                                   \\ \hline
Gender              & \begin{tabular}[c]{@{}l@{}}woman, female, girl, wife,\\ sister, daughter, mother\end{tabular}                                                  & \begin{tabular}[c]{@{}l@{}}man, male, boy, son,\\ father, husband, brother\end{tabular}                                                                                            \\ \hline
Race                & \begin{tabular}[c]{@{}l@{}}african, african american,\\ asian, black, hispanic, latin,\\ mexican, indian, \\ middle eastern, arab\end{tabular} & \begin{tabular}[c]{@{}l@{}}white, caucasian, european, \\ american, european, norwegian, \\ german, australian, english, \\ french, american, swedish,  \\canadian, dutch\end{tabular} \\ \hline
Sexual-orientation  & \begin{tabular}[c]{@{}l@{}}lesbian, gay, bisexual,\\ transgender, tran,\\ queer, lgbt,lgbtq,homosexual\end{tabular}                     & hetrosexual, cisgender                                                                                                                                                             \\ \hline
Religion            & \begin{tabular}[c]{@{}l@{}}jewish,buddhist,sikh,\\  taoist, muslim\end{tabular}                                                                & catholic, christian, protestant                                                                                                                                                    \\ \hline
Disability          & blind, deaf, paralyzed                                                                                                                         &                                                                                                                                                                                    \\ \hline
Social-class        & \begin{tabular}[c]{@{}l@{}}secretary, miner, worker, \\ machinist, nurse, hairstylist, \\ barber, janitor, farmer\end{tabular}                 & \begin{tabular}[c]{@{}l@{}}writer, designer, actor, \\ Officer, lawyer, artist,\\ programmer, doctor, \\ scientist, engineer, architect\end{tabular}                              \\ \hline
\end{tabular}}

    \caption{{The non-offensive identity (NOI) words used to describe the marginalized and non-marginalized groups in each sensitive attribute. For the disability-sensitive attributes, we use only words to describe disability due to the lack of words used to describe able-bodied.}}
    \label{tab:identity_groups}
\end{table*}
\end{document}